\pdfoutput=1

\documentclass[11pt]{article}

\usepackage[preprint]{acl}

\usepackage{times}
\usepackage{latexsym}

\usepackage{enumitem}
\setlist{noitemsep}

\usepackage[T1]{fontenc}

\makeatletter
\newcommand\footnoteref[1]{\protected@xdef\@thefnmark{\ref{#1}}\@footnotemark}
\makeatother

\usepackage[utf8]{inputenc}

\usepackage{microtype}

\usepackage{inconsolata}

\usepackage{graphicx}

\usepackage{listings}

\definecolor{background}{RGB}{255,255,245}
\definecolor{comment}{RGB}{0,128,0}
\definecolor{keyword}{RGB}{255,20,147}
\definecolor{class}{RGB}{0,0,225}
\definecolor{argument}{RGB}{65,105,225}
\definecolor{variable}{RGB}{220,20,60}  

\lstdefinelanguage{PromptoPython1}{
  language=Python,
  morekeywords=[2]{prompto,Settings,Experiment,ExperimentPipeline},
  sensitive=true,
  keywordstyle=[2]\color{class}\bfseries,
}

\lstdefinelanguage{PromptoPython2}{
  language=PromptoPython1,
  morekeywords=[3]{await},
  sensitive=true,
  keywordstyle=[3]\color{keyword},
}

\lstdefinelanguage{PromptoPython3}{
  language=PromptoPython2,
  morekeywords=[4]{data_folder,max_queries,max_attempts,parallel,file_name,settings},
  sensitive=true,
  keywordstyle=[4]\color{argument},
}

\lstdefinelanguage{PromptoPython}{
  language=PromptoPython3,
  morekeywords=[5]{experiment_settings,experiment,experiment_pipeline,process,run},
  sensitive=true,
  keywordstyle=[5]\color{variable},
}

\lstdefinestyle{mystyle}{
    language=PromptoPython,
    backgroundcolor=\color{background},
    basicstyle=\ttfamily\fontsize{8}{10}\selectfont,
    breaklines=true,
    captionpos=b,
    commentstyle=\color{comment},
    keywordstyle=\color{keyword},  
    stringstyle=\color{black},
    identifierstyle=\color{black},
    showstringspaces=false,
    numbers=none,
    frame=single,
    framesep=5pt,
    xleftmargin=10pt,
    xrightmargin=10pt,
    framexleftmargin=10pt,
    framexrightmargin=10pt,
    tabsize=4,
}

\usepackage{fancyvrb}

\title{\emph{Prompto}: An open source library\\ for asynchronous querying of LLM endpoints}

\author{Ryan Sze-Yin Chan$^{1}$, Federico Nanni$^{1}$, Angus R. Williams$^{2}$,\\\textbf{Edwin Brown$^{1,3}$, Liam Burke-Moore$^{2}$, Ed Chapman$^{1}$, Kate Onslow$^{2}$,}\\\textbf{Tvesha Sippy$^{2}$, Jonathan Bright$^{2}$, Evelina Gabasova$^{1}$}\\
$^1$ Research Engineering Group \& $^2$ Public Policy Programme (The Alan Turing Institute),\\ $^{3}$ Research Software Engineering Team (University of Sheffield)\\
 Corresponding authors: \tt\{rchan,fnanni\}@turing.ac.uk}

\begin{document}
\maketitle
\begin{abstract}
Recent surge in Large Language Model (LLM) availability has opened exciting avenues for research. However, efficiently interacting with these models presents a significant hurdle since LLMs often reside on proprietary or self-hosted API endpoints, each requiring custom code for interaction. Conducting comparative studies between different models can therefore be time-consuming and necessitate significant engineering effort, hindering research efficiency and reproducibility. To address these challenges, we present \texttt{prompto}, an open source Python library which facilitates asynchronous querying of LLM endpoints enabling researchers to interact with multiple LLMs concurrently, while maximising efficiency and utilising individual rate limits. Our library empowers researchers and developers to interact with LLMs more effectively and allowing faster experimentation, data generation and evaluation. \texttt{prompto} is released with an introductory video\footnote{\fontsize{8}{10}\url{https://youtu.be/lWN9hXBOLyQ}} under MIT License and is available via GitHub\footnote{\fontsize{8}{10}\url{https://github.com/alan-turing-institute/prompto}}.
\end{abstract}

\section{Introduction} \label{sec:introduction}



The field of Natural Language Processing is going through a massive transition since the introduction of Transformer-based Large Language Models (LLMs) \citep{vaswani2017attention, devlin-etal-2019-bert, radford2019language} which have demonstrated exceptional generalisation capability on a wide range of language-related tasks.

While user-friendly interfaces like ChatGPT, Gemini and Claude have made LLMs more accessible to the public, researchers nowadays increasingly interact with such models through programmatic interfaces and APIs. \citet{la2023language} noted several reproducibility issues with this \emph{Language-Models-as-a-Service (LMaaS)} paradigm \citep{sun2022black}, where language models are centrally hosted and typically provided on a subscription or pay-per-use basis (e.g., the OpenAI API\footnote{\fontsize{8}{10}\url{https://openai.com/api/}} and Google's Gemini API\footnote{\fontsize{8}{10}\url{https://ai.google.dev/gemini-api}}). To address some of them, \citet{biderman2024lessons} suggested a series of best practices when evaluating LLMs, such as sharing your exact prompts, parameter inputs and code and always providing model outputs.

In addition to LMaaS, there are now also several \emph{open} LLMs, defined here as those with broadly available model weights as in \citet{kapoor2024societal}, e.g. Llama \citep{meta2024introducing}, Gemma \citep{team2024gemma}, Aya \citep{aryabumi2024aya}. These "open" models are often accompanied by a \emph{model card} \citep{wolf-etal-2020-transformers} which provides instructions for executing them locally and for accessing their internals and weights. This allows users to create their own API endpoints for LLMs (e.g., via Ollama\footnote{\fontsize{8}{10}\url{https://ollama.com/}}) on their available hardware.

A significant challenge with both LMaaS and open LLMs is that conducting a comparative study across multiple models necessitates writing separate code to interact with each API and this obstacle further hinders already complex evaluation and reproducibility efforts. An additional problem is that APIs may have different constrains (e.g., \emph{query-per-minute (QPM)} rate limits), adding even more complexity to the engineering design. 

To address these limitations and simplify large-scale comparative studies across LLMs, we introduce \texttt{prompto}, an open source Python library for \emph{asynchronous} querying of LLM endpoints in a consistent and highly efficient manner. \texttt{prompto} uses asynchronous programming to efficiently interact with endpoints by allowing users to send multiple requests to different APIs concurrently. This eliminates idle wait times and maximises efficiency, especially when dealing with different rate limits. In contrast, in traditional synchronous programming, a user sends a single request to an endpoint and waits for a response from the API before sending another request, repeating for each query. \texttt{prompto} supports a range of LMaaS endpoints (e.g. OpenAI, Gemini, Anthropic) as well as self-hosted endpoints for querying local models (e.g. Ollama, Hugging Face's \texttt{text-generation-inference}\footnote{\fontsize{8}{10}\url{https://github.com/huggingface/text-generation-inference}} for serving models hosted on Hugging Face). The codebase is easily extensible to integrate new APIs and/or locally self-hosted models. For instance, we provide an example using Quart\footnote{\fontsize{8}{10}\url{https://github.com/pallets/quart}} in \texttt{prompto} to easily set up an endpoint for inferencing local models using \texttt{transformers} \citep{wolf-etal-2020-transformers}.

Inspired by \citet{biderman2024lessons}, the library promotes experiment reproducibility by facilitating the definition of all inputs/prompts within a single JSON Lines (JSONL) or CSV file. The file can encompass queries for various APIs and models, enabling parallel processing for even greater efficiency gains. The library's scalability allows it to handle large-scale experiments.  \texttt{prompto} also provides built-in functionalities for automatic evaluation of the obtained responses, such as allowing the user to apply scoring functions to model outputs, and model graded evaluation or LLM-as-a-judge \citep{zheng_judge_2023}.

In this paper, we present an overview of \texttt{prompto}, highlighting its modular design and flexibility. We present \texttt{prompto}'s functionalities, design choices, technical implementation and show its advantages in comparison with alternative approaches. We accompany its release with extensive documentation and a series of Jupyter Notebooks as tutorials, to allow the research community to easily explore all its functionalities. In addition to the library's provided examples\footnote{\fontsize{8}{10}Example usage of \texttt{prompto} can be found at \url{https://alan-turing-institute.github.io/prompto/examples/}\label{footnote:example}}, we provide an illustrative showcase of using our library and compare against a synchronous approach to query several LLM endpoints in parallel in Appendix \ref{sec:experiments}.

\section{Related Work and Motivation} \label{sec:related_work}

To support transparent and reproducible evaluation of large language models, a series of evaluation frameworks have been published in recent years, such as \emph{Language Model Evaluation Harness} (\texttt{lm-eval}\footnote{\fontsize{8}{10}\url{https://github.com/EleutherAI/lm-evaluation-harness}}) \citep{eval-harness, biderman2024lessons}, the UK AI Safety Institute's \emph{Inspect} framework (\texttt{inspect-ai}\footnote{\fontsize{8}{10}\url{https://github.com/UKGovernmentBEIS/inspect_ai}}), Confident AI's \emph{DeepEval} (\texttt{deeepeval}\footnote{\fontsize{8}{10}\url{https://github.com/confident-ai/deepeval}}), \emph{LLM Comparator} (\texttt{llm-comparator}\footnote{\fontsize{8}{10}\url{https://github.com/pair-code/llm-comparator}}). 

Nevertheless, conducting evaluation experiments on multiple large language models in order to compare their performance remains an engineering challenge, as each endpoint requires dedicated code to be written. To address this, our \texttt{prompto} library allows users to specify queries for various APIs and models within a single JSONL file, without having to interact directly with the details of each endpoint's infrastructure. This prioritises flexibility in experiment design, simplifies the setup for researchers and especially promotes reproducibility as all model parameters and prompts can be documented within a single file. After running an experiment, we record all model responses into an output experiment file which can be further analysed or used downstream in other pipelines.

Furthermore, our \texttt{prompto} library prioritises efficiency when running experiments. When interacting with LLMs, a sequential and synchronous approach where requests are sent to an endpoint and then waited upon for a response is highly inefficient, especially when dealing with different rate limits or when conducting comparisons across multiple models. \texttt{prompto} and tackles this challenge by leveraging asynchronous programming. 

The ability to interact with multiple LLMs concurrently and process requests asynchronously makes our library particularly well-suited for initial exploration and rapid comparison between different LLMs. Researchers can efficiently experiment with various models and gauge their performance on specific tasks. This might be particularly useful in experiments where an "expected" or "ideal" response is not known ahead of time since evaluation framework tools are often most useful for settings where there is pre-defined model response from which the performance of an LLM can be scored against (e.g. automated metrics such as BLEU \citep{papineni2002bleu}, HellaSwag \citep{zellers2019hellaswag} or MMLU \citep{hendrycksmeasuring}). \texttt{prompto} may also be preferred in settings where the responses are intended for human evaluation, for instance in \citet{rao2023assessing, rashkin-etal-2023-measuring, dash2023evaluation, williams2024large}.

Finally, \texttt{prompto}'s applicability goes beyond beyond evaluation tasks, and can be leveraged to generate synthetic datasets from LLMs for model training and development \citep{wang2022self, xu2024wizardlm, xu2024magpie}, or to create datasets specifically tailored for LLM distilation tasks, a technique for compressing knowledge from larger models into smaller, more efficient ones \citep{alpaca, wu2023lamini, peng2023instruction, gu2023knowledge}. 


\section{Library Design} \label{sec:library_design}

Our \texttt{prompto} library is an open source Python package\footnote{\fontsize{8}{10}\url{https://pypi.org/project/prompto/}} which facilitates processing of experiments of language models stored as JSONL files. The library has commands to process an experiment file to query models and store results. A user is able to query multiple models asynchronously and in parallel in a single experiment file.

We detail key components of setting up an experiment and core commands of the library. For a comprehensive exploration of \texttt{prompto}'s features, we kindly refer readers to the detailed documentation at \url{https://alan-turing-institute.github.io/prompto/}.

\subsection{Pipeline data folder} \label{subsec:data_folder}

\begin{figure}
    \centering
    \includegraphics[width=1.01\linewidth]{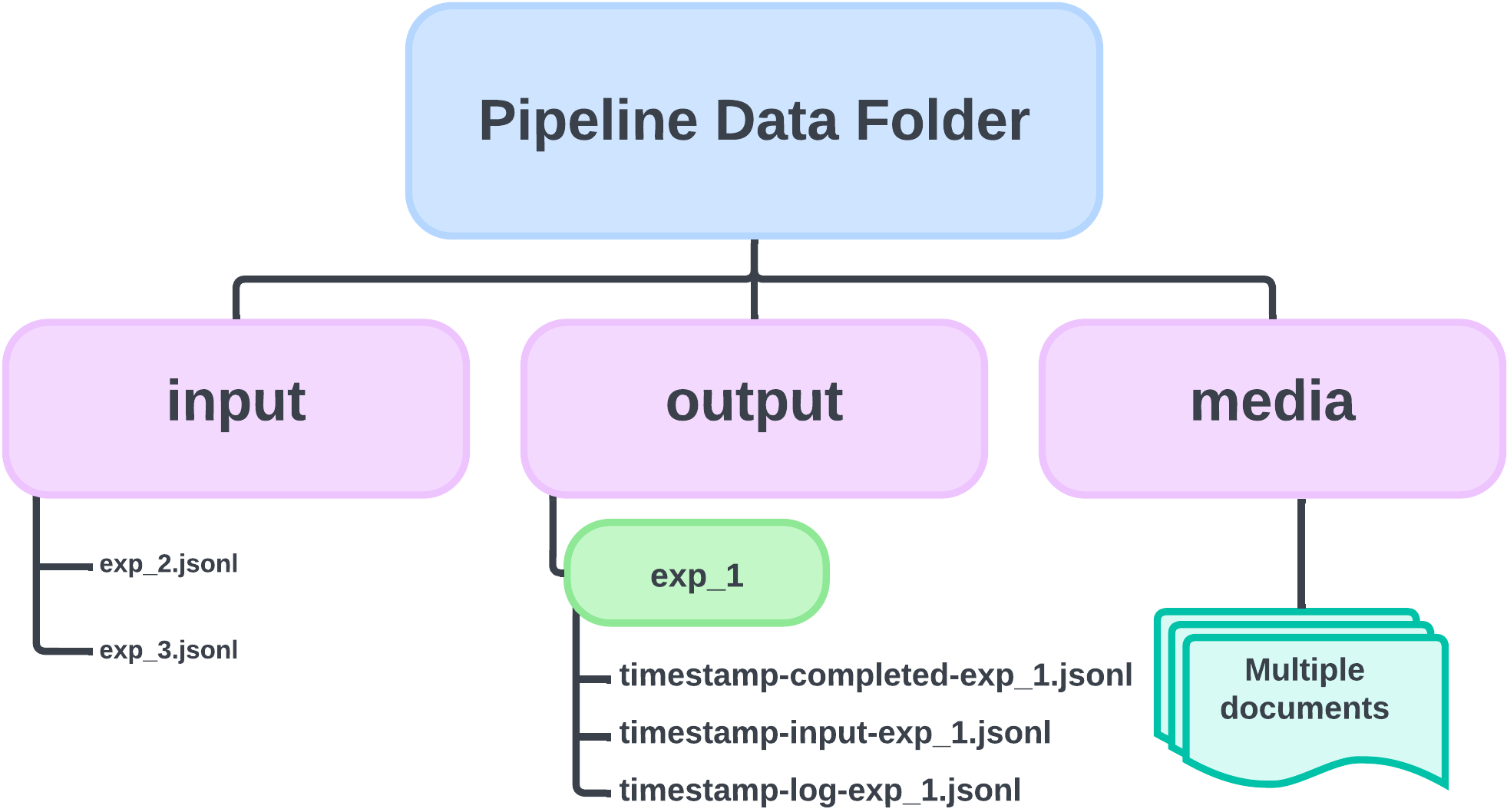}
    \caption{Illustration of the folder structure of the pipeline data folder after a \texttt{exp\_1.jsonl} experiment file has been processed while \texttt{exp\_2.jsonl} and \texttt{exp\_3.jsonl} are experiment files yet to be processed in the \texttt{input} folder. After processing \texttt{exp\_1.jsonl}, a \texttt{exp\_1} folder is created in the \texttt{output} folder which stores a completed experiment file, the original input file and a log file which are all timestamped to when the experiment was ran. The \texttt{media} folder stores any input media files for multimodal experiments.} 
    \label{fig:pipeline}
\end{figure}

For running an experiment in \texttt{prompto}, everything starts with setting up a \emph{pipeline data folder} (illustrated in Figure \ref{fig:pipeline}) which has several sub-folders:
\begin{itemize}
    \item \texttt{input}: contains input experiment JSONL files as described in \ref{subsec:experiment_file}
    \item \texttt{output}: contains results of experiment runs
    \item \texttt{media}: contains any input media files for multimodal model experiments
\end{itemize}

\noindent When an experiment is ran, a folder will be created in the \texttt{output} folder with the experiment name (the experiment file name without the "\texttt{.jsonl}" extension). We move the input file (originally is stored in the \texttt{input} folder) into the \texttt{output} folder and rename it to indicate it was the original input file to a particular run. The model responses are stored in a new "completed" JSONL file where each prompt dictionary will have a new "response" key storing the model response. A log file is also created to store any logs from the experiment run. Each of these files are timestamped to when the experiment started, allowing users to run multiple times the same experiment without interfering with the same output files and to track any changes to the input files between runs.  An illustration of this data folder can be found in Figure \ref{fig:pipeline}.

\subsection{Setting up a \texttt{prompto} experiment} \label{subsec:experiment_file}

A \texttt{prompto} \emph{experiment file} is a JSON Lines (JSONL) file which contains the prompts for an experiment along with any other parameters or metadata. Each line in the JSONL file is a valid JSON value\footnote{\fontsize{8}{10}In practice, this means each line is a collection of key/value pairs and in Python, this is realised as a dictionary.} which defines a particular input to a model. We refer to a single line in the JSONL file as a \emph{prompt dictionary} (or \texttt{prompt\_dict}) with keys:
\begin{itemize}
    \item \texttt{prompt}: the prompt to the model
    \item \texttt{api}: the name of the API to query (e.g. \texttt{"openai"}, \texttt{"gemini"}, etc.)
    \item \texttt{model\_name}: the name of the model to query (e.g. \texttt{"gpt-4o"}, \texttt{"gemini-1.5-flash"}, etc.)
\end{itemize}
The value of the \texttt{prompt} key is typically a \emph{string} type that is passed to the model to generate a response, but for certain APIs or models, this could take different forms. For instance, for some API endpoints like OpenAI and Gemini, this could be a \emph{list of strings} which we consider as a sequence of prompts to be sent as user messages in a chat interaction, or it could be a \emph{list of dictionaries} each with "role" and "content" keys which can be used to define a \emph{history} of a conversation. To foster the use of multimodal LLMs, \texttt{prompto} supports images and videos as inputs for several models. For instance, for the OpenAI API, it is possible to prompt with images by passing them through the "content" keys. These can be URL links to images on the web, or be stored locally in the \texttt{media} folder of the pipeline data folder (see Section \ref{subsec:data_folder})\footnote{\fontsize{8}{10}A full example with multimodal prompting in \texttt{prompto} with the OpenAI API can be found at \url{https://alan-turing-institute.github.io/prompto/examples/openai/openai-multimodal/}.}.

There are also optional keys that can be included in a prompt dictionary such as "\texttt{parameters}" which define the parameters or \emph{generation configuration} for a prompt such as the \emph{temperature}. Note that API services often have different names for the same generation parameter\footnote{\fontsize{8}{10}For example, to specify the maximum output tokens to generate, the OpenAI API uses \texttt{max\_tokens} while Gemini API uses \texttt{max\_output\_tokens}.}. In \texttt{prompto}, these parameters are not unified and the generation parameters specified in the prompt dictionary are directly passed to the API used. A "\texttt{group}" key could be defined to pass a user-specified grouping of the prompts which can be useful to process groups of prompts in parallel.

Note that CSV files can also be used as input where they get converted to JSONL to be processed. In this setting, the keys in the prompt dictionary discussed above are columns in the CSV file.

An experiment can be ran in the terminal using the \texttt{prompto\_run\_experiment} command line interface (CLI) by specifying the path to input experiment file with the \texttt{-{}-file} or \texttt{-f} argument. The user can also specify the path to the data folder (Section \ref{subsec:data_folder}) with the \texttt{-{}-data-folder} or \texttt{-d} argument\footnote{\fontsize{8}{10}By default, we assume the data folder is called \texttt{data/} in the current working directory.}:
\begin{Verbatim}[commandchars=\\\{\}]
  \textcolor{variable}{prompto_run_experiment} \\
    \textcolor{argument}{--file} path/to/experiment.jsonl \textbackslash \\
    \textcolor{argument}{--data-folder} data
\end{Verbatim}
For interacting with LLM endpoints, we often need to specify API keys or other variables. These can be set as environment variables, or they can be specified in a \texttt{.env} file\footnote{\fontsize{8}{10}By default, we look for a \texttt{.env} file in the current working directory, but a path can be specified using \texttt{-{}-env} or \texttt{-e}.} as key-value pairs:
\begin{Verbatim}
  OPENAI_API_KEY=...
  GEMINI_API_KEY=...
\end{Verbatim}
The \texttt{prompto\_run\_experiment} CLI then starts asynchronously sending requests for each prompt dictionary which includes details of the input, the API to send to and the model name to query. There are several other arguments to the command for providing flexibility of how this process occurs\footnote{\fontsize{8}{10}Full details of each of these arguments can be found in our documentation at \url{https://alan-turing-institute.github.io/prompto/docs/pipeline/}} such as the maximum queries to send per minute (\texttt{-{}-max-queries} or \texttt{-mq}), the maximum number of retries if errors (such as rate limit or connection errors) occur (\texttt{-{}-max-attempts} or \texttt{-ma}). To adhere to strict API rate limits, we use the \texttt{max-queries} (per minute) argument to determine how frequently to send our requests asynchronously. If \texttt{max-queries} is set to $10$, we simply send our requests every $60/10=6$ seconds. Since we use an asynchronous approach, we do not need to wait for a response before sending another request, ensuring that prompts are sent at correct intervals. To handle any errors (e.g. API failures or timeouts), we use the \texttt{max-attempts} argument to determine the maximum number of attempts for a prompt. In the case of an unexpected error, prompts are added to the back of the queue and are retried a maximum number of times. If the maximum is reached, we return and log the error in the response.

For example, to run an experiment where we asynchronously send 50 queries per minute (one query every 1.2 seconds) with 5 maximum retries, we can simply run the following command:
\begin{Verbatim}[commandchars=\\\{\}]
  \textcolor{variable}{prompto_run_experiment} \textbackslash \\
    \textcolor{argument}{--file} path/to/experiment.jsonl \textbackslash \\
    \textcolor{argument}{--data-folder} data \textbackslash \\
    \textcolor{argument}{--max-queries} 50 \textbackslash \\
    \textcolor{argument}{--max-attempts} 5
\end{Verbatim}
Furthermore, \texttt{prompto} supports sending requests to different APIs or models in parallel (\texttt{-{}-parallel} or \texttt{-p}). Our implementation of the "parallel" processing again uses asynchronous programming allowing multiple queues of prompts to be managed concurrently within a single thread of execution rather than utilising multiple CPU cores and so is more resource efficient. If parallel processing of different APIs or models is requested, the user can fully customise how queues of prompts are constructed and set different rate limits for each queue. This is particularly useful when querying multiple API endpoints with different rate limits\footnote{\fontsize{8}{10}Full details of parallel processing can be found at \url{https://alan-turing-institute.github.io/prompto/docs/rate_limits/}}.

Users can use all of \texttt{prompto}'s functionalities directly in Python rather than using the CLI. For instance, to run an experiment in Python:
\begin{lstlisting}[language=PromptoPython]
from prompto import Settings, Experiment

experiment_settings = Settings(
    data_folder="data",
    max_queries=50,
    max_attempts=5,
)
experiment = Experiment(
    file_name="path/to/experiment.jsonl",
    settings=experiment_settings,
)

await experiment.process()
\end{lstlisting}
In this example code, we are utilising the \texttt{Settings} and \texttt{Experiment} classes of the \texttt{prompto} library. The \texttt{Settings} class defines the settings of the experiment and stores all the paths to the relevant data folders, whereas the \texttt{Experiment} class stores all the relevant information for a single experiment and takes in a \texttt{Settings} object during initialisation. The \texttt{Experiment} class has an async method called \texttt{process} which runs the experiment.

For a more illustrative walkthrough of running experiments, the documentation includes examples of typical workflows with associated notebooks\footnoteref{footnote:example}.

\subsection{The \texttt{prompto} pipeline} \label{subsec:experiment_pipeline}

Our \texttt{prompto} library also has the functionality to run a pipeline which continually looks for new experiment JSONL files in the input folder using the \texttt{prompto\_run\_pipeline} command which takes in the same settings arguments as the \texttt{prompto\_run\_experiment} command:
\begin{Verbatim}[commandchars=\\\{\}]
  \textcolor{variable}{prompto_run_pipeline} \textbackslash \\
    \textcolor{argument}{--data-folder} data \textbackslash \\
    \textcolor{argument}{--max-queries} 50 \textbackslash \\
    \textcolor{argument}{--max-attempts} 5
\end{Verbatim}
This command initialises the process of continually checking the input folder for new experiments to process. Consequently, there is no need to pass in a path to an experiment file to process as with \texttt{prompto\_run\_experiment}. If a new experiment is found, it is processed and the results and logs are stored in the output folder as explained in Section \ref{subsec:experiment_run}. The pipeline will continue to check for new experiments until the process is stopped. In the case where there are several experiments in the input folder, the pipeline will process the experiments in the order that the files were last modified.

In Python, it is possible to initialise this process in the following way:
\begin{lstlisting}[language=PromptoPython]
from prompto import Settings, ExperimentPipeline

experiment_settings = Settings(
    data_folder="data",
    max_queries=50,
    max_attempts=5,
)
experiment_pipeline = ExperimentPipeline(
    settings=experiment_settings,
)

experiment_pipeline.run()
\end{lstlisting}


\subsection{Automatic evaluation} \label{subsec:auto_eval}

A common use case for \texttt{prompto} is to evaluate different models, where we first need to obtain a large number of responses and then subsequently evaluate those responses. \texttt{prompto} provides functionality to automatically evaluate model responses.

One approach is to apply a simple \emph{scoring function} to responses. A scoring function is typically lightweight such as performing string matching to some \emph{target} output or some regex pattern matching. We have some built in scoring functions in the library such as a \texttt{match()} which determines whether the model response matches some expected response pre-defined by the user, or \texttt{includes()} which determines if the model responses includes a substring which is defined by the user. 

In future work, we hope to expand this functionality to more advanced and computationally expensive scoring functions such toxicity classifiers \citep{inan2023llama, Detoxify}. While these can already be implemented as custom scorers in the library, scorers are applied independently to each output meaning batching is not fully utilised in the current implementation.

To automatically apply scoring functions to model outputs during an experiment run, one can simply use the \texttt{-{}-scorers} argument to specify a comma-separated list of scoring functions:
\begin{Verbatim}[commandchars=\\\{\}]
  \textcolor{variable}{prompto_run_experiment} \textbackslash \\
    \textcolor{argument}{--file} path/to/experiment.jsonl \textbackslash \\
    \textcolor{argument}{--data-folder} data \textbackslash \\
    \textcolor{argument}{--scorers} match,includes
\end{Verbatim}

\noindent Another common approach to automatic evaluation is using an LLM to judge the outputs of a model \citep{zheng_judge_2023}. To perform an LLM-as-judge evaluation, the user must provide a prompt template which will be used to generate prompts to the Judge LLM for evaluation. In \texttt{prompto}, we treat this evaluation as simply as another \texttt{prompto} experiment where we obtain a new set of prompts using some judge evaluation template which includes a model response. An LLM-as-judge evaluation can be performed automatically after running an experiment. Full details of running LLM-as-judge evaluations and all other evaluations features available with \texttt{prompto} can be found at \url{https://alan-turing-institute.github.io/prompto/docs/evaluation/}.


\subsection{Rephrasing prompts} \label{subsec:rephrasing_prompts}

It is often useful to be able to rephrase/paraphrase a given prompt, particularly in evaluation settings since performance of models can vary significantly with choice of prompt or wording \citep{gonen-etal-2023-demystifying, gsm-symbolic, hughes2024best}. In \texttt{prompto}, we provide functionality to simply use another language model to rephrase a given prompt. A \emph{rephrasal experiment} can be constructed by using a prompt template to prompt a model to rephrase/paraphrase. The responses from the rephrasal experiment (along with the original prompts), can be sent to another model to obtain responses for later evaluation or for any other purpose. Full details and some examples of utilising the rephrasing functionality in \texttt{prompto} can be found at \url{https://alan-turing-institute.github.io/prompto/docs/rephrasals/}.


\subsection{Adding new APIs and models to \texttt{prompto}} \label{subsec:adding_new_api}

We have designed \texttt{prompto} to be easily extensible to integrate new LLM API endpoints and models. In particular, a user can add a new API by creating a new class which inherits from the \texttt{AsyncAPI} class from the \texttt{prompto} library. The user must then implement an async method \texttt{query} which asynchronously interacts with the model endpoint. Full details on implementing a new model endpoint can be found at \url{https://alan-turing-institute.github.io/prompto/docs/add_new_api/}.

\section{Conclusions and Future Work} \label{sec:conclusion}

We present \texttt{prompto}, an open source library designed to facilitate researchers to efficiently query large language models which reside on proprietary or self-hosted API endpoints. Using \texttt{prompto} is simple and all inputs/prompts can be easily defined in a single JSONL file, which can encompass prompts for various APIs and models. Outputs are then stored in output JSONL files allowing researchers to easily share experiment outputs promoting reproducibility. Consequently, our library enables faster experimentation and evaluation. \texttt{prompto} is an ongoing open source project and we welcome contributions from the community to ensure support for a wide range of LLM endpoints and new features. 

There are a number of interesting avenues for extending \texttt{prompto}. In the future, we hope to continue focusing on extending the evaluation pipelines discussed in Section \ref{subsec:auto_eval}. For evaluation, integration of larger-scale NLP pipelines can be useful in order to efficiently apply more advanced scoring functions such as toxicity classifiers and other safety risk classifiers for automatic safety evaluation. Furthermore, integrating prompto with data visualisation libraries can bring additional plotting features to the library to enable deeper exploration of model performance.

Another direction of work is to extend the rephrasing pipelines in \texttt{prompto} (see Section \ref{subsec:rephrasing_prompts}). A particular area of interest is extending the pipeline for translation of prompts. This can be useful for synthetic data generation applications. The current pipeline enables the use of LLMs for translation which have demonstrated remarkable potential in handling multilingual machine translation (MMT) \citep{zhu-etal-2024-multilingual}, however extending this pipeline to dedicated MMT models, such as NLLB \citep{costa2022no,nllb2024scaling}, could particularly be effective for translating to lower resourced languages. Moreover, a rephrasing pipeline can also be important in evaluation settings to generate a wider set of prompts to test a model. However, further research is required in order to do this effectively.

Lastly, future work includes expanding our tool's accessibility by developing direct local model integration capabilities. The current implementation requires setting up an API endpoint for querying, but we envision simplifying this process by enabling direct model instantiation (e.g., using \texttt{transformers} \citep{wolf-etal-2020-transformers}). While toolkits for deploying and serving LLMs, such as \texttt{text-generation-inference} or \texttt{vLLM} \citep{kwon2023efficient}, already simplify setting up efficient endpoints, this approach would reduce overhead for researchers who wish to query local models and utilise offline batched inference, allowing for more streamlined experimentation and rapid prototyping.



\section*{Acknowledgments} \label{sec:acknowledgements}

We thank Yi-Ling Chung, Florence Enock, Kobi Hackenburg for useful discussions on this library. This work was partially supported by the Ecosystem Leadership Award under the EPSRC Grant EPX03870X1, The AI Safety Institute \& The Alan Turing Institute.

\bibliography{acl_latex}

\appendix

\section{Experiments} \label{sec:experiments}

We provide some illustrative examples\footnote{\fontsize{8}{10}\label{footnote:hardware}All experiments were ran on a 2021 MacBook Pro with M1 Pro and 32 GB of memory.} of using our \texttt{prompto} library and compare against a traditional synchronous approach to querying LLM endpoints. We show that our our pipeline provides significant speedups, especially when querying different models either on separate LLM endpoints (Appendix \ref{subsec:experiment_2}) or available on the same API service (Appendix \ref{subsec:experiment_3}).

In each of the following examples, we consider the task of generating instruction-following data using the Self-Instruct approach of \citet{wang2022self} and \citet{alpaca}. We take a sample of $100$ prompts from the instruction-following data\footnote{\fontsize{8}{10}\url{https://github.com/tatsu-lab/stanford_alpaca/blob/main/alpaca_data.json}} from \citet{alpaca} and apply the same prompt template. We then use these as prompt inputs to different models using \texttt{prompto}. Full details of how we obtained a sample of prompts along with notebooks to run each of the following experiments can be found at \url{https://alan-turing-institute.github.io/prompto/examples/system-demo/}. 

Further examples including examples with prompting multimodal models with images and videos can be found on our GitHub repo and documentation\footnoteref{footnote:example}.


\subsection{Querying LLM endpoints asynchronously vs synchronously} \label{subsec:experiment_1}

We first compare \texttt{prompto} against a synchronous approach for three different LLM endpoints: the OpenAI API for \texttt{gpt-3.5-turbo}, Gemini API for \texttt{gemini-1.5-flash} and Ollama API for Llama 3. For using \texttt{prompto} for OpenAI and Gemini APIs, we send requests at 500 queries per minute (QPM). For Ollama, we send only at $50$ QPM due to the limitations of the machine which we are running the Ollama server from\footnoteref{footnote:hardware}. We report run-times of the two approaches for obtaining $100$ responses to sample prompts in Table \ref{tab:experiment_1}. 

For OpenAI and Gemini, even with a small sample of $100$, we observe a significant speed up (of around 9 times or 12 times, respectively). Additionally, note that for OpenAI and Gemini APIs, there are tiers for which the query per minute rate limit is much higher than 500 QPM so further gains could be reached. For Ollama, we only observe a modest improvement in run-time since the Ollama API handles async requests by queuing the prompts for computation, so requests still get completed one-at-a-time. Therefore, in this setting \texttt{prompto} is simply putting prompts in a queue for the Ollama server.

\begin{table}[ht]
    \centering
    \small
    \caption{Run-time in seconds to obtain responses from 100 prompts from each API using a synchronous approach and using \texttt{prompto}.}
    \label{tab:experiment_1}
      \begin{tabular}{@{}lccc@{}}
        & \textbf{OpenAI} & \textbf{Gemini} & \textbf{Ollama} \\
        \hline
        sync & 126.31 & 163.49 & 271.45 \\
        \texttt{prompto} & 13.92 & 14.09 & 268.59 \\
        \hline
        \textbf{speedup} & 9.07 & 11.60 & 1.01 \\
    \end{tabular}
\end{table}

\subsection{Querying different LLM endpoints in parallel vs synchronously} \label{subsec:experiment_2}

As mentioned in Section \ref{subsec:experiment_run}, \texttt{prompto} supports sending requests to different APIs in parallel for greater efficiency gains. Further, the user can easily specify how many queries to send to each API or model per minute. To illustrate, we consider the same APIs and models as in Section \ref{subsec:experiment_1}, but query the models in parallel. We compare against the baseline synchronous approach which we expect to be close to the sum of the individual run-times of the synchronous approaches in Table \ref{tab:experiment_1}. We report run-times of the two approaches to obtain a total of $300$ prompts ($100$ to each model/API) in Table \ref{tab:experiment_2}.

We observe a $2$ times speedup when using \texttt{prompto} with parallel processing. The \texttt{prompto} run-time is close to how long it took to process the Ollama requests in Section \ref{subsec:experiment_1} which is expected since the run-time of querying different APIs or models in parallel is simply dictated by the slowest API or model to query.

\begin{table}[ht]
    \centering
    \small
    \caption{Run-time in seconds to obtain responses from 100 prompts from each API in one run using a synchronous approach and enabling parallel processing of APIs with \texttt{prompto}.}
    \label{tab:experiment_2}
    \begin{tabular}{@{}lc@{}}
         & \textbf{Overall} \\
        \hline
        sync & 558.74 \\
        \texttt{prompto} & 269.06 \\
        \hline
        \textbf{speedup} & 2.08 \\
    \end{tabular}
\end{table}

\subsection{Querying different models from the same endpoint in parallel vs synchronously} \label{subsec:experiment_3}

Lastly, we illustrate how a user can query from different models available at the same endpoint using \texttt{prompto}. For this experiment, we consider the OpenAI API to query from three different models: \texttt{gpt-3.5-turbo}, \texttt{gpt-4} and \texttt{gpt-4o}. As with the previous two experiments, we will compare the run-times of querying the models individually as well as using \texttt{prompto} with parallel processing (i.e. sending requests to each model in parallel). For each model, we send requests at a rate of 500 QPM. We report run-times of this experiment in Table \ref{tab:experiment_3}.

For each model and with parallel processing of the models, using \texttt{prompto} offers a signficant speedup in obtaining responses. Focusing on the synchronous run-times, we can see GPT-4o and GPT-4 are slower to obtain responses for than GPT-3.5-Turbo. Comparing using \texttt{prompto} with parallel processing against synchronously querying each model, we obtain an approximately $35$ times speedup with \texttt{prompto}. As with the previous experiment in Section \ref{subsec:experiment_2}, the run-time of \texttt{prompto} with parallel processing is roughly the time to query the slowest model, GPT-4 in this case.

\begin{table}[ht]
    \centering
    \small
    \caption{Run-time in seconds to obtain responses from 100 prompts from different models from the OpenAI API using a synchronous approach and using \texttt{prompto} and enabling parallel processing of models.}
    \label{tab:experiment_3}
    \begin{tabular}{@{}lcccc@{}}
         & \textbf{GPT-3.5} & \textbf{GPT-4} & \textbf{GPT-4o} & \textbf{Overall} \\
        \hline
        sync & 130.73 & 392.21 & 241.24 & 705.38 \\
        \texttt{prompto} & 14.29 & 19.79 & 18.11 & 19.30 \\
        \hline
        \textbf{speedup} & 9.15 & 19.82 & 13.32 & 36.55 \\
    \end{tabular}
\end{table}



    
    


\end{document}